\documentclass[10pt,twocolumn]{article}

\usepackage[margin=0.72in,columnsep=0.24in]{geometry}
\usepackage[T1]{fontenc}
\usepackage{newtxtext,newtxmath}
\usepackage{microtype}
\usepackage{graphicx}
\usepackage{booktabs}
\usepackage{array}
\usepackage{tabularx}
\usepackage{xcolor}
\usepackage{hyperref}
\usepackage{enumitem}
\usepackage[font=small,labelfont=bf]{caption}
\usepackage{fancyhdr}
\usepackage{titlesec}
\usepackage{amsmath}
\usepackage{float}

\definecolor{ink}{HTML}{18202A}
\definecolor{accent}{HTML}{315C8C}
\definecolor{rulegray}{HTML}{D8DCE2}
\hypersetup{colorlinks=true,linkcolor=accent,citecolor=accent,urlcolor=accent,
  pdftitle={SlopBench: How Well Can We Rank Language Models by Slop?},
  pdfauthor={Dhruv Roongta, Harsha Gaddipati, Anh Tuan Huynh},
  pdfsubject={A multi-domain benchmark of repetitive AI writing across 18 language models},
  pdfkeywords={AI slop, benchmark, language models, text generation, evaluation, writing quality}}
\setlist{nosep,leftmargin=1.25em}
\titleformat{\section}{\large\bfseries\color{ink}}{\thesection}{0.55em}{}
\titleformat{\subsection}{\normalsize\bfseries\color{ink}}{\thesubsection}{0.5em}{}
\titlespacing*{\section}{0pt}{1.25em}{0.45em}
\titlespacing*{\subsection}{0pt}{0.9em}{0.3em}
\newcommand{\score}{\textsc{SlopScore}}
\newcommand{\repo}{\href{https://github.com/hgaddipati1118/slop-index}{github.com/hgaddipati1118/slop-index}}

\title{\vspace{-1.1em}\textbf{SlopBench: How Well Can We Rank Language Models by Slop?}\\
\large A Multi-Domain Benchmark of Repetitive AI Writing}
\author{
Dhruv Roongta$^{1,*}$ \quad Harsha Gaddipati$^{1,*}$ \quad Anh Tuan Huynh$^{1,\dagger}$\\[-0.1em]
\small $^1$Slashy, San Francisco, California, USA\\[-0.2em]
\small $^*$Equal contribution. \quad $^\dagger$Work conducted while interning at Slashy.
}
\date{}

\begin{document}
\maketitle
\thispagestyle{fancy}

\begin{abstract}
SlopBench asks which models produce the stiff, repetitive prose readers call AI slop, a question detectors leave open once they have classified a text as machine-written. We evaluated eighteen models on 112 hand-written tasks in email, social posts, essays, and workplace chat, sampling each model on each task up to ten times, for 19,928 outputs in all. SlopBench scores four surface behaviors a reader can check by hand: length against the word band each task specifies, opener repetition across a model's own samples of one task, and paragraph rhythm and fixed lexical constructions against pre-ChatGPT human corpora. Under one fixed weighting, Kimi K2.6 scores lowest at 21.1 and Mistral Large highest at 40.6. Across 500 random reweightings Kimi has the lowest score in 58 percent of draws and Mistral the highest in 97 percent. No draw preserves the full order of the eighteen, and a scenario bootstrap leaves exactly one of those ranks unambiguous. We ran three further checks on that middle order: a crowd arena, an AI detector, and lexical diversity. None of them confirmed the order. We therefore report the four behaviors separately and treat the composite as one weighting among many, and we release the prompts, outputs, reference statistics, and scoring code.
\end{abstract}

\section{Introduction}

Most model evaluations ask whether an answer is correct, safe, or preferred. A response can pass all three tests and still sound mass-produced, with a slow opening, a reused frame, uniform paragraphs, and stock phrasing. Readers often call this writing ``AI slop.''

We ask which models are least likely to write this way. AI detection cannot answer that on its own, because an authorship label does not separate model-written texts by how they read. To compare models we need signals that vary among them and that a reader can inspect.

The term slop still lacks an agreed definition, and the two studies closest to the question operationalize it differently. Shaib et al. derive a taxonomy from a survey of 19 practitioners in writing, journalism, linguistics, NLP, and philosophy, and report pairwise Cohen's $\kappa$ between $-0.15$ and $0.29$ on binary slop labels, alongside Gwet's AC1, a prevalence-robust statistic, of 0.12 to 0.42 \cite{shaib2026slop}, while Miklian and Katsos show that the features separating AI from human text do not necessarily predict which human texts readers accuse of being AI-written \cite{miklian2026accusations}. Authorship, style, quality, and authenticity overlap in practice, yet a detector that settles authorship leaves the other three open.

SlopBench covers \emph{surface slop}: the repeated or inflated patterns that can be counted directly in model output. Truth, reasoning, coherence, and usefulness are outside its scope. Models complete ordinary writing tasks up to ten times. Each axis has its own reference. We score paragraph rhythm and lexical constructions against human writing from the same broad medium, opener repetition against the model's other samples of the same task, and length against a target band the task itself specifies.

The paper contributes:
\begin{enumerate}
    \item A public set of 19,928 outputs from 18 models on 112 writing tasks in four domains.
    \item Four interpretable measurements of length, opener repetition, paragraph rhythm, and fixed lexical constructions, the last two anchored to statistics from pre-ChatGPT human corpora.
    \item A model and domain scorecard with bootstrap rank intervals, which resolve one endpoint model and leave the middle of the table unordered.
    \item Three negative results that limit what the score can claim: a detector that saturates, a crowd comparison too noisy to interpret, and a lexical-diversity axis that points the wrong way.
\end{enumerate}

Under the composite formula Kimi K2.6 sits at the clean end and Mistral Large at the other. Only Mistral Large's rank survives a scenario bootstrap intact, while every other position, Kimi's included, spans a range. Changing the writing domain or the metric weights changes the order again. The crowd ranking is too noisy to confirm or contradict the overall order. The practical question is therefore which model is least sloppy for a given kind of writing.

\section{Related Work}

\subsection{Slop, style, and linguistic diversity}

Shaib et al. provide the closest formal treatment of AI slop \cite{shaib2026slop}, with a taxonomy covering information utility, information quality, and style quality, annotated at both the document and span level. Agreement on the binary label is low enough that the authors report a prevalence-corrected statistic alongside $\kappa$, and the need for both measures already suggests that slop is not one settled quantity. Their study asks what readers mean by slop. We measure how often current models produce a small set of surface signs, and only within style quality.

Two studies measure linguistic convergence at scale, each against a different comparison point. Sourati et al. analyze 880,000 texts and find that writing assistants reduce variance in writing complexity and weaken author-specific signals, measuring what a tool removes from a person's own draft \cite{sourati2026shrinking}, while Gude et al. report lower syntactic and, especially, lexical diversity in instruction-tuned models than in the earlier non-instruction-tuned generation, comparing one model generation against another \cite{gude2026diversity}. Both stop short of comparing a model against a human corpus. Section~\ref{sec:diversity} finds the opposite sign when we do: every model is more diverse than its reference. Both document a narrowing of variation, though neither tests repeated attempts at the same task. We therefore sample repeatedly, since one varied answer says little about whether a model returns to the same forms across its remaining samples.

Shaib et al. separately find favored part-of-speech templates in generated text, at rates that vary by task and model \cite{shaib2024templates}. SlopBench uses a simpler exact-opener measure, which is easy to reproduce but misses paraphrased and syntactic templates.

\subsection{Benchmarks for subjective writing}

LitBench tests creative-writing evaluators with large preference datasets and a held-out online study \cite{fein2026litbench}. HoWToBench breaks writing quality into a tree of subfeatures instead of asking one judge for an overall score \cite{feng2026howtobench}. We follow HoWToBench in publishing the parts alongside the composite rather than a verdict on its own, with Section~\ref{sec:human} giving the limits of our own human comparison.

Human rankings depend on the voters, sometimes to a degree that undercuts the aggregate. Garbacea et al. fit per-user Bradley-Terry models for 115 arena voters and find a mean correlation of $\rho=0.04$ with the aggregate ranking, with 57 percent showing near-zero or negative correlation \cite{garbacea2026personalized}. Their per-user Elo rankings track the aggregate more closely, at $\rho=0.43$, and Elo is the rating system our own arena uses. Chatbot Arena uses blind pairwise comparisons, large vote counts, Bradley-Terry estimation, and uncertainty intervals \cite{chiang2024arena}. Our arena also hides model names, but it has fewer votes and uses sequential Elo, so we treat it as a limited validity check rather than a controlled annotation study.

\subsection{What good benchmarks make explicit}

HELM argues for multi-metric evaluation and direct reporting of tradeoffs \cite{liang2023helm}. $\tau^2$-bench specifies its tasks, controls task generation, verifies end states, and studies errors \cite{barres2026tau2}. SlopBench follows most of that practice. Section~\ref{sec:measurement} defines the measured behaviors, and Sections~\ref{sec:diversity}, \ref{sec:detector}, and~\ref{sec:human} report where the evaluator fails. We also release the raw outputs and sample every task up to ten times per model.

AI-text detectors solve a different task, and RAID finds their performance brittle across domains, decoding strategies, models, and attacks \cite{dugan2024raid}. We do not evaluate Pangram as a detector but ask whether its score can rank style among texts already known to be AI-written. In our sample every text receives the same score, so it cannot.

\section{SlopBench}

\subsection{Task suite}

SlopBench contains 112 English-language scenarios: 30 emails, 27 essays, 27 workplace chats, and 28 social posts. Each scenario specifies a writer, recipient, context, goal, and desired length. Tasks include declining a discount, posting a short launch note, reporting a delay, thanking a vendor, writing a cover letter, and replying to a disputed claim.

All prompts are hand-written, fictional, and modelled on routine writing requests, and each targets a failure mode we wanted to stress. Some demand a short answer, while others forbid a formal greeting, a list, an engagement prompt, or an abstract moral. Across all 19,928 released records, every model sees the same prompt for a given scenario and one system message for that domain.

Scenario files include a \texttt{split} field that marks 74 prompts as public and 38 as held out. In this release the label has no analytical effect: we commit the held-out prompts in plain text alongside the public ones, and no prompt is withheld from any reported score. We keep the field because a future run should withhold those 38, but the numbers here come with no contamination guarantee. Appendix~\ref{app:schema} gives the full composition by category.

The intended matrix is ten generations for each model and scenario, and the release contains 19,928 of 20,160 expected outputs, or 98.85 percent (Table~\ref{tab:data}). Missing outputs remain missing and none are imputed.

\begin{table}[t]
\centering
\caption{SlopBench at a glance. The arena row reports the number of recorded games. Votes are a separate figure, and Section~\ref{sec:arena} explains why the two differ.}
\label{tab:data}
\small
\begin{tabular}{lr}
\toprule
Item & Count \\
\midrule
Models & 18 \\
Scenarios & 112 \\
Domains & 4 \\
Samples per model and scenario & up to 10 \\
Expected outputs & 20,160 \\
Observed outputs & 19,928 \\
Completion rate & 98.85\% \\
Arena games (18 models) & 2,914 \\
\bottomrule
\end{tabular}
\end{table}

\subsection{Models and generation}

Eighteen models from eleven developers are covered: Alibaba, Anthropic, DeepSeek, Google, Meta, MiniMax, Mistral, Moonshot, OpenAI, xAI, and Zhipu. We use the model names stored with the outputs, since that is what the scorer reads. Each record also stores the route used to reach the model, prompt and completion token counts, reasoning tokens when available, latency, and estimated cost.

Eleven models were called through their developer's own API. Six came through resellers: Fireworks for DeepSeek V4 Pro, GLM 5.2, and Kimi K2.6, and OpenRouter for MiniMax M3, Mistral Large, and Qwen 3.7 Max. The remaining model, Muse Spark 1.1, came through an OpenAI-compatible custom endpoint rather than Meta's own API (Table~\ref{tab:models}). Both models at the ends of the headline ranking were reached through non-first-party endpoints: Kimi K2.6, the lowest, via Fireworks, and Mistral Large, the highest, via OpenRouter. Resellers can differ from the first-party endpoint in quantization, default sampling parameters, and system-prompt handling. We did not audit those differences. Two display names also omit version details that the records retain: Gemini 3.1 Pro is the \texttt{preview} build, and Mistral Large is the dated \texttt{2512} snapshot.

Generation used each route's default sampling configuration. The benchmark therefore measures the served model together with whatever defaults that route applies. The merged release has no run-level manifest for timestamps, temperature, top-p, or seeds. Raw records preserve the output metadata, but the stochastic sample cannot be regenerated exactly. Numerical results hold for the released output file and repository commit. The same model name at another setting or date may behave differently.

\begin{table}[tb]
\centering
\caption{How each model was reached, taken from \texttt{usage.model\_id} in the released records. Seven of eighteen did not come through the developer's own API: six through a reseller, including the models at both ends of the headline ranking, and Muse Spark 1.1 through an OpenAI-compatible custom endpoint.}
\label{tab:models}
\small
\begin{tabular}{lrl}
\toprule
Developer & Models & Route \\
\midrule
Anthropic & 4 & Anthropic \\
OpenAI & 4 & OpenAI \\
Google & 2 & Google \\
xAI & 1 & xAI \\
\midrule
Meta & 1 & OpenAI-compatible \\
DeepSeek & 1 & Fireworks \\
Moonshot & 1 & Fireworks \\
Zhipu & 1 & Fireworks \\
Alibaba & 1 & OpenRouter \\
MiniMax & 1 & OpenRouter \\
Mistral & 1 & OpenRouter \\
\bottomrule
\end{tabular}
\end{table}

\subsection{Human reference corpora}

Each domain has exactly one human reference corpus, all of it written before ChatGPT: EnronSent for email (50,046 sent messages, 5.98M words, public domain), the Blog Authorship Corpus for essays (60,000 posts, 22.4M words, research use only), Sentiment140 for social posts (120,000 tweets, 1.63M words), and DISCO, a corpus of open-source project Discord messages, for workplace chat (250,000 turns, 3.51M words). Because licenses differ, the repository publishes derived statistics and rebuild scripts, while the raw texts stay with their original sources.

These corpora provide reference values for two of the four axes: mean paragraph-length variation anchors rhythm, and pooled construction rates anchor lexical tells. The length band and the opener measure draw on none of them. Because these are domain-level norms and not matched human responses to the 112 prompts, they cannot show what a person would have written for any one task. We describe the two anchored axes as measuring \emph{distance from a human domain reference}, which is a narrower claim than distance from human writing in general.

\subsection{Data governance}
\label{sec:governance}

No Slashy customer data was used: the benchmark contains no customer email bodies, prompts, account records, or private customer metadata, its prompts are fictional, and every evaluated text was generated for the project. Public reference corpora do contain writing by real people, including historical Enron employees. For that reason we use aggregate statistics under the source licenses and do not redistribute the corpora.

The arena records model-pair results, scenario identifiers, coarse country codes, dwell time, and a salted hash of the visitor address used for rate limiting. It records no Slashy customer content. We report only aggregate vote counts and model ratings.

\section{Measurement}
\label{sec:measurement}

\subsection{Operational definition}

SlopBench measures four axes over the samples a model produces for a scenario: length inflation $C$, repeated openers $T$, paragraph rhythm flattening $R$, and lexical tells $L$. Each axis maps to $[0,100]$, where larger values mean more of the measured behavior. The score describes behavior only, and it is not a probability that a text is bad or AI-written.

\paragraph{Length inflation.}
\label{sec:length}
Each scenario has a target word-count interval with upper bound $u_s$. Answers below the band receive no penalty. Per-output inflation is $c(x)=\max(0,\,n(x)/u_s-1)$ for word count $n(x)$, and a model's score in a domain averages those ratios before clipping:
\begin{equation}
\label{eq:length}
C=100\min\!\left(1,\ \frac{1}{|X|}\sum_{x\in X}c(x)\right)
\end{equation}
over that model's outputs $X$. A model that always stops at the upper bound scores 0, and one whose outputs average double the bound scores 100.

Clipping the average instead of each output leaves the axis sensitive to its own tail. DeepSeek V4 Pro shows the size of that effect. Its median output is 28 percent \emph{under} the band ceiling, and only 2.6 percent of its outputs exceed the bound by more than double. Three replies run 2,150 to 2,454 words for a prompt asking for 25 to 70. They push its axis score to 19.2, the highest in the set, and widen its bootstrap rank interval to ten ranks. Clipping each output before averaging would place it at 8.3 instead. The axis is measuring a tail: three replies out of 1,120, against a median that sits well inside the band.

\paragraph{Repeated openers.}
For each model and scenario, we lowercase the first five word tokens of every sample. Let $p_{m,s}$ be the share of the $N_{m,s}$ samples that use the most common exact opener. A model that never repeats an opener still yields $p_{m,s}=1/N_{m,s}$, so we subtract that floor. The scorer averages $p_{m,s}$ over the scenarios with at least two samples, then applies the correction once using the mean sample count $\bar N_m$:
\begin{equation}
\label{eq:opener}
T_m=100\min\!\left(1,\ \max\!\left(0,\ \frac{\bar p_m-1/\bar N_m}{1-1/\bar N_m}\right)\right).
\end{equation}
Correcting per scenario and then averaging is a different operation, but $N_{m,s}=10$ in 1,990 of the 1,995 populated cells, so the two give identical scores to one decimal place for all 18 models. Exact matching is transparent but conservative: ``Thanks for sending this over'' and ``Thank you for sending this'' count as different openers. Section~\ref{sec:opener-robust} reports what happens when we strip salutations and markdown headings before taking the window.

\paragraph{Paragraph rhythm.}
\label{sec:rhythm}
For each output with at least two paragraphs we compute the population standard deviation of paragraph word counts. Let $v_h$ be the mean of this statistic in the human domain reference and $v_m$ the model mean. Rhythm flattening is
\begin{equation}
\label{eq:rhythm}
R_m=100\min\left(1,\max\left(0,1-\frac{v_m}{v_h}\right)\right).
\end{equation}
Only in email is the axis informative. With no recoverable paragraph structure in the Blog Authorship Corpus or Sentiment140, $v_h$ is undefined for essays and social posts. Workplace chat differs: paragraph lengths in DISCO turns vary little ($v_h=3.59$ words), and every model varies more. Equation~\ref{eq:rhythm} therefore clips to zero for all 18 models.

An axis is dropped only where the baseline cannot define it. Essays and social posts have no $v_h$, so they lose the rhythm term and the weights renormalize over the remaining three. Workplace chat does have a $v_h$. There the scorer computes the axis, it returns zero for all 18 models, and it keeps its 0.20 weight. The rhythm term therefore takes a fifth of the chat weight while separating none of the models there. Every workplace-chat score is 3.4 to 8.0 points below its renormalized counterpart, and each overall score about 1.3 points below. The effect is close to proportional, changing the level of the chat column more than its order. In Figure~\ref{fig:axes} and Table~\ref{tab:full} the rhythm column averages the email score with that structural zero, halving the email score. Workplace-chat scores do not sit on the same scale as essay and social scores.

\paragraph{Lexical tells.}
The scorer counts seventeen fixed constructions per 1,000 words, listed in full in Appendix~\ref{app:tells}, but excludes the general dash counter from scoring so dashes do not count twice. Each tell scores 0 at or below its rate in the human corpus, then rises linearly to 100 at three times that rate. A tell absent from the human reference receives 50 when it appears, after which the axis averages the available tell scores.

Pilot tests left this list unchanged but removed whole candidate axes, starting with compression ratio, whose confidence intervals overlapped across models. Lexical diversity (MTLD) separated the models cleanly but pointed the wrong way at every scale we checked. Each of the 18 models has \emph{higher} MTLD than its matching human corpus in every domain, 72 model-domain pairs in all (Section~\ref{sec:diversity}). It cannot work as an axis that reads lower lexical diversity as more slop. Most of the classic markers sit at a median of zero. In all four domains the median score across models stays at or below the human rate for \emph{delve}, \emph{tapestry}, ``as an AI,'' ``do not hesitate,'' ``hope this email finds you,'' and the engagement prompts, so those classes sit at a median of zero and dilute the axis. Only em-dash use shows a large gap in every corpus. Human use is near zero in all four corpora. It runs from 0.000 per 1,000 words in EnronSent and Sentiment140 to 0.026 in the Blog Authorship Corpus and 0.041 in DISCO, whose 2019--2020 writers could type the character and largely did not. Model rates average 1.5 to 16.5 per 1,000 words by model, and span 0.5 to 26.7 across model-domain pairs. Even the lowest model-domain rate is twelve times the highest human rate. Against the two corpora with no em dashes at all the ratio is undefined. The gap may reflect typography and corpus age as well as style, which makes the em dash a corpus-relative signal and not a general rule.

In practice the axis is much coarser than the linear rule suggests. Of the 64 domain-by-tell combinations, 33 sit at zero, 15 saturate at 100, 12 drop out because neither the corpus nor the models use the construction, three take the flat 50, and exactly one lands strictly between the endpoints. It mostly counts which constructions a model overuses rather than the degree of overuse. Scored classes vary by domain, from 11 in social to 14 in email.

\subsection{Fixed scalar score}
\label{sec:scalar}

The released scorer uses
\begin{equation}
\score=0.35C+0.30T+0.20R+0.15L,
\end{equation}
with weights renormalized over the axes the domain baseline can define. Averaging the four domain scores without further weighting gives the overall figure. The authors set the axis weights and did not learn them from human judgments.

The companion website shows a different number. The two numbers are far apart, and the following paragraph gives the reason. The website min-max normalizes each axis across whatever models are currently on the board, then blends 40 percent crowd Elo with the four mechanical axes at 15 percent each. Its score is board-relative. Zero there means ``cleanest model presently listed'' rather than ``matches the human corpus.'' Adding a model can rescale the others, so the same system scores 24.6 on an 18-model board and 21.8 on the 39-model board now live. After that rescaling crowd Elo enters at 40 percent, giving crowd Elo the largest weight in the blend. Across the same 18 models, the rankings the website and this paper produce have a Spearman correlation of $0.21$: the website's sloppiest model is Gemini 3.5 Flash, which this paper places tenth, and its cleanest is Muse Spark 1.1, which this paper places seventeenth. We report the anchored score throughout, with Section~\ref{sec:human} comparing it against the crowd ratings while the website figure stays outside that comparison.

\subsection{Blind human comparisons}
\label{sec:arena}

The companion arena shows two outputs for the same scenario. It hides the model names, randomizes which side each appears on, and asks the visitor to pick the sloppier one or mark both. Proof-of-human checks, a reading-time floor, and rate limits filter obvious abuse. Ratings use sequential Elo with $K=32$, and a higher model rating means its outputs were flagged less often and looked cleaner.

We report only what the snapshot lets us verify: the 18 models in this paper have 142 to 181 recorded games each, 2,914 in total. The arena's separate counter reports 1,587 accepted votes: 1,209 chose one output, and 378 marked both. Ties count as draws and still credit a game to each side, so those votes imply 3,174 model-games, 260 more than the per-model counts contain. Two mechanisms could account for the difference: the arena serves a larger roster than this benchmark run, and it caps votes per visitor at aggregation time. With the raw vote log unpublished, we could not rebuild the ratings to check. Every arena-based score in this paper rests on the 2,914 model-games, and we do not treat the 1,587 figure as a sample size.

Older votes do not retain the exact output identifiers. We can compare models but cannot link a vote to features of a specific output. We use the arena as exploratory validation, which falls short of a completed human study.

\section{Research Questions}
\label{sec:rq}

We ask five questions of the released run. Three further subsections fall outside them: Section~\ref{sec:opener-robust} tests the axis with the widest spread, Section~\ref{sec:diversity} reports a rejected axis, and Section~\ref{sec:repro} covers reproduction and missing data.

\begin{enumerate}[label=\textbf{RQ\arabic*.},leftmargin=2.6em]
    \item Which models receive the lowest mechanical slop score?
    \item How much does the answer change across writing domains?
    \item Which axes drive each model's score?
    \item Can a leading AI detector rank degrees of slop?
    \item Do mechanical rankings agree with blind human comparisons?
\end{enumerate}

We recompute all four axes from the released raw output file. Running the repository's scoring command reproduces the committed score log exactly. To estimate rank uncertainty, we resample scenarios within each domain 500 times, recompute every score, and report the 2.5th to 97.5th percentiles of each model's rank distribution. To test weight sensitivity, we draw 500 weight vectors, sampling each axis weight uniformly from 0.10 to 0.45 and normalizing to sum to one.

Those draws hold three other constants fixed. A lexical tell saturates at three times the human rate, a tell whose human rate is zero receives 50, and the overall mean weights the four domains equally. Section~\ref{sec:ranking} reports what happens when we vary each one.

The published board scores each domain separately over the axes that domain supports, then averages the four. The sensitivity analysis instead applies a single normalization across all domain-axis pairs. Email and workplace chat each carry four scored axes against three for essays and social posts, so those two domains count for more in the sensitivity form. Even under the published weights the two orderings place four models differently, so part of the reported instability comes from how the two methods define the aggregate. We rechecked the result against a baseline computed the same way as the trials. Under either version none of the draws preserves the full order, so the ordering stays unstable.

We compare the frozen mechanical scores with a live arena snapshot at the model level. Pearson correlation measures linear association and Spearman correlation compares ranks. We give Fisher-$z$ intervals over the 18 models, and those intervals agree with a nonparametric bootstrap to within 0.05. Those intervals describe sampling variation across the 18 models. They cannot propagate the error in each Elo point estimate, which Section~\ref{sec:human} shows is large, because the arena publishes only rounded sequential ratings and no per-vote output pairs. The comparison neither supports an association nor establishes the absence of one.

\section{Results}

\subsection{Which models produce the least measured slop?}
\label{sec:ranking}

\begin{figure}[tb]
    \centering
    \includegraphics[width=\columnwidth]{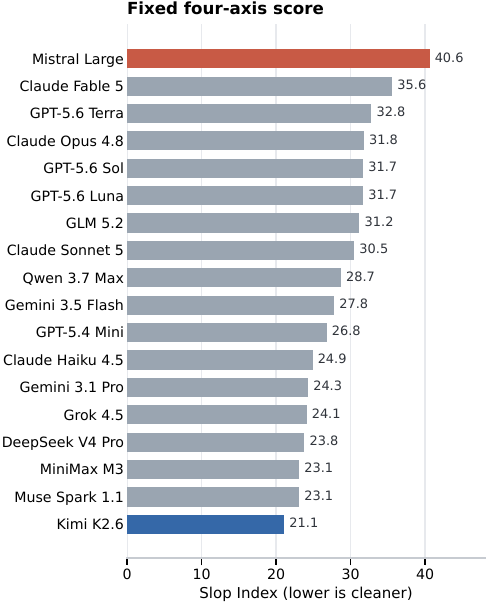}
    \caption{Slop Index for all 18 models. Lower is cleaner, and 0 is the floor on every axis: no length inflation, no repeated openers, and no distance from the human domain reference on rhythm and lexical tells. Section~\ref{sec:ranking} gives the rank intervals: only Mistral Large's rank is resolved to a point, and the bootstrap intervals of the two lowest-scoring models overlap.}
    \label{fig:ranking}
\end{figure}

Figure~\ref{fig:ranking} shows the fixed ranking, with Kimi K2.6 lowest at 21.1 and MiniMax M3 and Muse Spark 1.1 following at 23.1. Mistral Large scores highest at 40.6, followed by Claude Fable 5 at 35.6.

Ranks run sloppiest first throughout this paper: rank 1 is the worst score and rank 18 the cleanest. Across 500 alternative weight draws Mistral Large stays at rank 1 in 97 percent of trials and Kimi K2.6 at rank 18 in 58 percent. No draw preserves the full order. A scenario-level bootstrap then gives 95 percent rank intervals for every model, reported in Table~\ref{tab:full}. Only Mistral Large's interval covers a single rank, while Kimi K2.6 spans 16 to 18 and four models have intervals that include rank 18. Treating overlapping intervals as ties leaves four groups, with six models in the cleanest group, so the data do not support a claim such as ``model A is exactly third best.''

Among the formula constants, sensitivity is concentrated in the axis weights. Moving the tell ceiling from three times the human rate to 2, 4, 6, 8, or 10 moves no model by more than two positions. Replacing the constant 50 for a tell with no human baseline by 0, 25, 75, or 100 does the same, as does dropping such tells entirely. Weighting the four domains by their scenario counts instead of equally swaps a single adjacent pair, MiniMax M3 and Muse Spark 1.1. Kimi K2.6 is the lowest-scoring model under all eleven variants. The ranking is sensitive to the relative weight of the four axes, while the constants used inside the axis formulas move it very little.

\subsection{The answer changes by domain}

\begin{figure}[tb]
    \centering
    \includegraphics[width=\columnwidth]{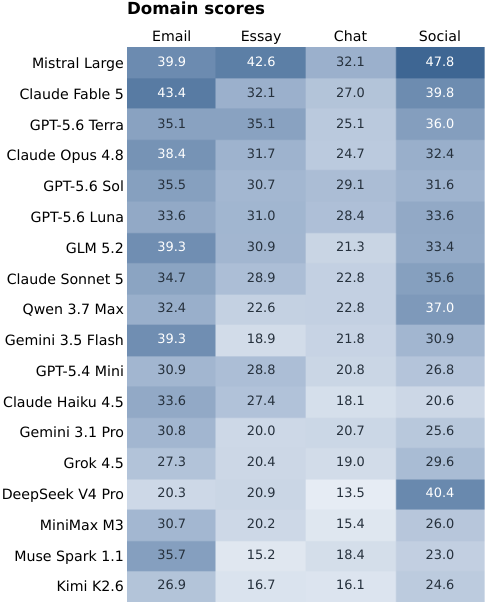}
    \caption{Domain scores under the fixed formula, shaded from 10 (lightest) to 50 (darkest). Darker means more measured slop. Rows follow the overall score, sloppiest first.}
    \label{fig:domains}
\end{figure}

Rankings change sharply by domain (Figure~\ref{fig:domains}). DeepSeek V4 Pro scores 13.5 in workplace chat and 40.4 on social posts, and Gemini 3.5 Flash scores 18.9 on essays and 39.3 on email. DeepSeek V4 Pro wins email (20.3) and workplace chat (13.5); Muse Spark 1.1 wins essays (15.2) and Claude Haiku 4.5 wins social posts (20.6), while Kimi K2.6 stays near the clean end everywhere without winning a single domain.

Only email separates models on all four axes. Essays and social posts have no paragraph-variance baseline, and in workplace chat the rhythm term is zero for every model. Essays and social posts therefore drop rhythm and renormalize the weights, while workplace chat keeps the term (Section~\ref{sec:rhythm}). The four columns therefore use three different scales. Email gives the most complete single-domain result. The four columns show whether the ordering holds from one domain to another.

\subsection{Models fail in different ways}

\begin{figure}[tb]
    \centering
    \includegraphics[width=\columnwidth]{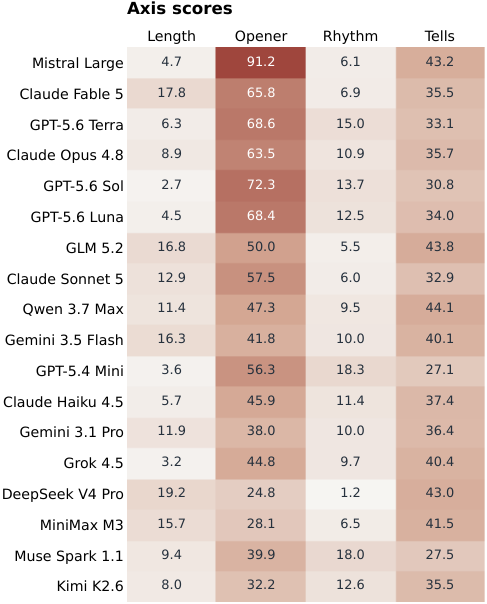}
    \caption{Mean axis scores, shaded from 0 (lightest) to 95 (darkest). Darker means more of that measured behavior. Rhythm is email only, the one domain where the axis separates models (Section~\ref{sec:rhythm}).}
    \label{fig:axes}
\end{figure}

Similar total scores can come from different behavior (Figure~\ref{fig:axes}). Repeated openers account for most of Mistral Large's score, at 91.2 against a median of 48.7, while Claude Fable 5 inflates length more than nearby models at 17.8. DeepSeek V4 Pro repeats openers least of any model at 24.8 and flattens rhythm least at 1.2, yet it has the highest length-inflation score at 19.2. GPT-5.4 Mini is near the low-inflation end at 3.6 but flattens rhythm most at 18.3.

Repeated openers have by far the widest spread, ranging from 24.8 to 91.2, against spreads of 17.1 for rhythm, 17.0 for tells, and 16.5 for length. Every model's median output sits below its band ceiling, so the axis responds to a tail of long outputs rather than to habitual padding, as DeepSeek V4 Pro shows below. Because the profiles differ this much, changing axis weights readily reorders the middle of the table.

\subsection{Robustness checks on the opener axis}
\label{sec:opener-robust}

Repeated openers take 0.30 of the weight and show nearly four times the spread of the next widest axis, 66.4 against 17.1, giving them more influence on the published order than any other axis. Salutations are one possible inflator. In 474 of the 536 email model-scenario cells the most common opener is a greeting addressed to the recipient the prompt names, such as ``hi carol thanks for reaching.'' The axis counts a repeated opening even when addressing a named recipient the same way is correct email practice. A further 5.7 percent of outputs begin with a markdown heading, which the five-token window captures instead of the prose. Re-scoring with leading headings and salutations removed lowers every model by 1.2 to 8.7 points. The order barely changes. Nine models shift, eight by one position and Gemini 3.1 Pro by two, and every one stays inside its bootstrap tie group. Dropping only from 91.2 to 90.0, Mistral Large stays an outlier because its repetition goes well beyond a shared greeting.

Opener repetition also varies with the prompt, and scenario identity accounts for 39 percent of the explained variance. Some prompts constrain the opening hard: 78 percent of all 18 models open \texttt{email.followup.001} with the same five tokens, against a median of 20 percent across scenarios. Splitting the variance in within-scenario repetition gives 0.0229 between models and 0.0148 between scenarios. Of the explained variance, 61 percent is attributable to the model and 39 percent to the prompt. Most of the explained variance comes from the model, but a different scenario mix would change the values. That is why the rank intervals in Table~\ref{tab:full} come from a scenario-level cluster bootstrap and not an output-level one.

\subsection{Models are more lexically diverse than the corpora}
\label{sec:diversity}

We computed MTLD \cite{mccarthy2010mtld} over 120 sampled outputs per model and domain. Every model exceeds the mean MTLD of its matching human corpus in every domain, 72 model-domain pairs in all. That gap is widest where the human texts are shortest. Some of it may be a length artifact. MTLD needs 20 tokens, and the Sentiment140 and DISCO messages are short enough that only the longer ones get scored. Email gives the closest comparison, because the reported lengths sit in the same range. EnronSent messages have a median of 90 words and a mean of 120, and the email task bands have a median upper bound of 110. Every model still exceeds the corpus there, from 107.3 for GPT-5.4 Mini to 149.8 for Grok 4.5 against 101.5 for the corpus.

Lexical diversity fails as an axis here, and the result makes no claim about writing quality. A model can reach for rarer words and still produce the repetition the other axes catch. Mistral Large does exactly that, sitting mid-pack on diversity while scoring 91.2 on repeated openers. The result does undercut the common claim that ``AI writing is lexically impoverished'' when that claim names no reference corpus and does not control for length.

\subsection{The detector saturates}
\label{sec:detector}

We sent 18 texts, spanning 65 to 544 words, from six of these models to Pangram's browser scorer (version 3.3.2). All 18 were labeled 100 percent AI-generated with high confidence, with no variance at all. Every input was model-written, so the detector's classification was never in question. The sample tests whether its score varies inside that class, and it does not. We cannot say more from a sample this size. A later pass covering more models fell outside the pinned release, so we do not report its numbers here.

In this sample the detector gives the same authorship label to model writing at both ends of the measured range. Its scores may vary by accident, and that variation would need its own validation before anyone could use it to measure slop.

\subsection{Mechanical scores and crowd Elo show little observed association}
\label{sec:human}

\begin{figure}[t]
    \centering
    \includegraphics[width=\columnwidth]{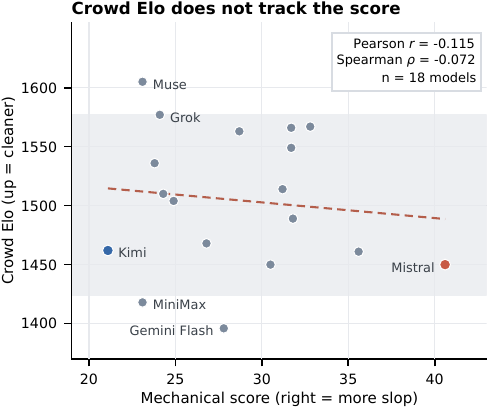}
    \caption{Model-level comparison between the mechanical score and crowd Elo. If both measures captured the same thing, the plot would show a downward slope. The dashed fit is nearly flat. The shaded band covers the central 90 percent of a single model's rating when all 18 are simulated as equally sloppy under the snapshot's own parameters. Fifteen of the 18 observed ratings fall inside it, one exactly on its upper bound. Kimi K2.6 and Mistral Large are colored as in Figure~\ref{fig:ranking}.}
    \label{fig:human}
\end{figure}

Figure~\ref{fig:human} compares the fixed mechanical score with crowd Elo, where we expect a negative relationship because higher mechanical scores mean more measured slop while higher Elo means voters judged the writing cleaner. We observe Pearson $r=-0.115$ (95 percent CI $[-0.55,+0.37]$, $p=0.65$) and Spearman $\rho=-0.072$ (95 percent CI $[-0.52,+0.41]$, $p=0.78$), treating the 18 models as the sampling unit. Both intervals run from a moderate negative association to a moderate positive one. The study cannot tell whether that association is moderately negative, negligible, or moderately positive.

The crowd ratings are too imprecise to support a strong reading of the correlation. We simulated the snapshot under the null that all 18 models are equally sloppy, using its own parameters: 2,914 model-games, $K=32$, the observed 23.8 percent draw rate, and uniform pairing. The central 90 percent of a single model's rating then falls between 1424 and 1577. Fifteen of the 18 observed ratings fall inside that interval, one of them exactly on its upper bound. Their 209-point spread sits at the 87th percentile of the null spread. The null simulation alone, with no difference in slop between models, reproduces most of the vertical extent of Figure~\ref{fig:human}.

A near-zero correlation is therefore unsurprising given how imprecise the estimate is at this sample size. On its own it does not establish that the mechanical axes miss what voters notice. Voters may judge quality, authenticity, or taste despite the wording of the question, and aggregate ratings can hide large differences among voters \cite{garbacea2026personalized}. Separating those explanations needs a larger balanced study at the output level, and this snapshot is too small for that.

The companion site publishes a score that puts 40 percent of the weight on the crowd Elo and spreads the other 60 percent evenly over the four mechanical axes. With a Spearman correlation this close to zero, that blend is not validated as a single construct. It averages two rankings not shown to measure the same thing, so we report them separately.

\subsection{Reproducibility and missingness}
\label{sec:repro}

Running the public scorer on the released JSONL reproduces the committed mechanical score log exactly, and every mechanical score in this paper comes from that log. Output keys contain no duplicates. The matrix is missing 232 outputs, or 1.15 percent: 140 from Muse Spark 1.1, 77 from GPT-5.6 Luna, and 15 from GLM 5.2. Scoring averages available data and does not penalize missing calls. The gaps cluster. GPT-5.6 Luna is missing 77 of its 270 workplace-chat outputs, including six of the 27 chat scenarios entirely, and Muse Spark 1.1 is missing five essay scenarios entirely, so those domain scores rest on a smaller scenario set than the rest of the table. We also rescored using only the scenarios where all eighteen models are complete. Both ends of the ranking stay unchanged, and seven models move within the middle, each staying inside its published rank interval.

The result file, scorer, scenarios, baseline statistics, and baseline rebuild scripts are public \cite{slopbench2026repo}, but exact regeneration is not possible without a run manifest and pinned software environment. Future benchmark runs will record output hashes, generation dates, provider parameters, and failed calls.

\section{Discussion}

\subsection{Read the leaderboard as a scorecard}

Under the released formula Kimi K2.6 has the lowest score and Mistral Large the highest. The data do not support reading that pair as a verdict on the whole board. No model wins everywhere: DeepSeek V4 Pro takes two domains, Muse Spark 1.1 and Claude Haiku 4.5 one each. The bootstrap pins down only one of the eighteen ranks.

Open-ended writing has no single ground truth. LitBench and HoWToBench break quality into preferences or subfeatures \cite{fein2026litbench,feng2026howtobench}. SlopBench is most useful as a compact scorecard: one summary number beside the domain and axis results that produced it.

\subsection{Slop is not AI detection}

AI authorship, measurable style, and perceived slop are different targets: a model can be easy to detect yet pleasant to read, or avoid every marker on a fixed list and still read as generic. Human surface style can also sit on top of a poor answer. Saturated Pangram scores are consistent with treating AI authorship, measurable style, and perceived slop as three distinct targets, while our crowd comparison is too noisy to give evidence either way.

SlopBench does not identify whether a person used AI, only how outputs from known models compare on a fixed task suite. The benchmark is meant for model selection and analysis. Academic-misconduct detection, hiring decisions, and authorship attribution fall outside what it supports.

\subsection{Slop belongs to a model version and setup}

Post-training, system prompts, decoding, and provider defaults all affect style, so each score belongs to a model version under one setup rather than to a model family or company. Benchmark runs should be dated and preserved. A provider that changes the served model should trigger a rerun.

Repeated sampling matters because one polished output does not reveal whether a model returns to the same opener on its next nine attempts. The design measures a distribution rather than one selected answer.

\section{Limitations and Future Work}
\label{sec:limits}

\paragraph{Construct coverage.}
Our four axes cover surface form only, omitting relevance, factuality, coherence, voice fit, and information density, all of which appear in broader definitions of slop \cite{shaib2026slop}. Here slop refers only to the measured surface forms.

\paragraph{Human references.}
No reference corpus contains matched responses to the benchmark prompts: Enron email is old workplace writing, Sentiment140 reflects a 2009 character limit, DISCO contains Discord messages rather than Slack, and the Blog Authorship Corpus is not a controlled essay sample. These sources give historical base rates but cannot establish a human ceiling.

\paragraph{Metric validity.}
\label{sec:metric}
Exact five-word matching misses syntactic paraphrases. Our tell axis compares a per-output mean model rate against a pooled corpus rate. These are different statistics, with the model side giving short outputs more weight. Rhythm separates models only in email. Scenario length bands were set by the authors, although matched human answers would have given a stronger empirical basis.

Length inflation clips the average of the per-output ratios instead of each ratio, leaving the axis sensitive to a model's longest few outputs (Section~\ref{sec:length}). In workplace chat rhythm returns zero for every model but keeps its 0.20 weight. Chat scores therefore sit on a different scale from essay and social scores (Section~\ref{sec:rhythm}). The weight-sensitivity analysis aggregates domains differently from the released scorer, so some of the reported instability is a matter of definition and not of data (Section~\ref{sec:rq}). A future version should bring length inflation and workplace-chat rhythm into line with the axis definitions, and run the sensitivity analysis on the same aggregation the released scorer uses. We report the released behavior so that the numbers here match the published scorer exactly.

\paragraph{Human study.}
Voters were self-selected visitors to a public website. They were unpaid, gave no demographic information, and were neither recruited nor screened. The site retains only a salted address hash, a coarse country code, and dwell time. No institutional review was sought. None of the texts under comparison was written by a person. The arena data form an opportunity sample rather than a purpose-built annotation study: the historical votes lack output identifiers, and the ratings lack confidence intervals. Section~\ref{sec:human} shows that at 2,914 games the ratings give little evidence against a null in which every model is equally sloppy.

More voting would not fix this by itself. With $K$ held fixed, sequential Elo settles at a spread set by $K$ rather than by the number of games, so more games do not narrow it \cite{cortez2026elo}. Simulating the null at ten and at thirty times the current volume leaves the median spread at 170 to 173 Elo, essentially unchanged. To resolve the models, replace the sequential update with a batch Bradley-Terry fit, whose standard error at the current 162 games per model is near 27 Elo. Precision then improves with vote volume: about 4,500 balanced comparisons across the board give 16 Elo, and 18,000 give 8 Elo. To make the comparison against the mechanical score interpretable, add \emph{models}. The Fisher-$z$ interval treats each model's score as fixed, so its width depends on how many systems are on the board and not on how many votes each receives. On the Fisher-$z$ scale the 95 percent half-width is 0.51 at $n=18$. Reaching 0.30 needs about 46 models, and reaching 0.25 about 64. A stronger study would also retain output identifiers, balance the comparisons, include scenario effects, and use a small expert overlap set to test whether voters read the instructions the same way.

\paragraph{Model setup.}
Provider defaults differ. Our merged run lacks a complete manifest. Future runs should record temperature, top-p, seed support, reasoning mode, API date, and provider model identifier. Each model should have the same number of successful samples, or missingness should enter the uncertainty estimate.

\paragraph{Contamination and reuse.}
Because every prompt is public, any model trained after this release may have seen the suite. The \texttt{split} field marks 38 scenarios as held out, but they are published and scored like the rest, so the present numbers give no protection against contamination. Because twenty scenarios also served in the pilot that chose which axes to keep, part of the evaluation set informed the measurement design. The next run can fix both by withholding the held-out prompts and drawing pilot scenarios from outside the scored set.

\paragraph{English and culture.}
All tasks and references are in English. Judgments about directness, formality, and repetition depend on culture and context. Other languages require new prompts, references, and annotators.

The existing model outputs can be reused in a subsequent validation. It should:
\begin{enumerate}
    \item Pre-register the axes, the weights, and the primary analysis, and publish them from a repository that preserves its full commit history. Publish a versioned manifest, a data card, output hashes, and a one-command build for every table and figure.
    \item Replace sequential Elo with a batch Bradley-Terry fit and collect roughly 4,500 balanced, output-level comparisons with retained identifiers, which brings the per-model standard error near 16 Elo. Increase the board toward 46 models before interpreting any correlation with the mechanical score.
    \item Add lexical and syntactic template measures while retaining exact openers as an interpretable baseline.
\end{enumerate}

\section{Conclusion}

SlopBench compares surface slop across 19,928 outputs from 18 models, and under the released weighting Kimi K2.6 records the lowest mechanical score and Mistral Large the highest. Only Mistral Large's rank is resolved to a point, with the bootstrap leaving the rest in four overlapping groups. The ranking changes again with the writing domain and the axis weights.

None of the three supported the ranking: the crowd comparison and the detector were inconclusive, and lexical diversity failed as an axis. A commercial detector gave all 18 texts we sent it the same score, so it separated nothing. A null simulation with no real differences largely reproduces the spread in the crowd ratings from 2,914 recorded model-games. That comparison is too coarse at this volume to count as evidence either way. And lexical diversity, which earlier work reports falling in model-assisted and instruction-tuned text, is higher than in the pre-ChatGPT corpus in all 72 pairs we tested. Each result narrows what a slop score can claim.

The result is better read as a scorecard than as a single index. Models differ across four visible behaviors, so a single number hides the behavior on which a given model scores worst. Separate scores make the ranking easier to inspect and to challenge. Released outputs and code let others test a different definition against the same samples.

\section*{Competing Interests}

All three authors are affiliated with Slashy, a commercial company that builds an AI-assisted email client. That affiliation creates two competing interests relevant to the interpretation of this work. The company sells a product in one of the four domains scored here. Email also has the most scenarios and is the only domain where all four axes separate models, making its results the most complete set of single-domain results in the paper. The company also selects third-party models for that product, so the ranking bears on purchasing decisions the authors' employer makes.

We mean this work to be checked: the prompts, the generated outputs, the scoring code, and the derived statistics are public, and every mechanical score here comes from running the published scorer unmodified. The scored models include the ones the company uses. The roster of 18 is the one that existed when the run was frozen. Later additions are visible in the repository. We report three negative results. Section~\ref{sec:scalar} shows that the score defined here disagrees with the company's own public leaderboard. No provider funded, reviewed, or was consulted about this work.

This study was not pre-registered. The public repository is a squashed mirror of a private one. The axes, weights, and harness all arrive in a single commit. Their history cannot be used to show that they were fixed before the results were seen. The weights are author-chosen, with the sensitivity analysis in Section~\ref{sec:ranking} the only check on them. Section~\ref{sec:limits} lists pre-registration as the first thing the next run should add.

\section*{Ethics and Use of AI Assistance}

Section~\ref{sec:governance} covers the data in full: no Slashy customer data was used, every scored text was generated by a model for this project, and the reference corpora are represented only through aggregate statistics. Section~\ref{sec:limits} describes the arena participants. They were unpaid, anonymous, self-selected visitors, and no institutional review was sought.

We used AI assistance for literature search, prose editing, LaTeX, and figure code. An author recomputed every numeric claim in this paper from the pinned artifacts, with no number taken from a model, and each related-work claim was checked against the cited paper. We are responsible for all content.

\section*{Data and Licensing}

We are releasing the 112 prompts, the 19,928 generated outputs, and the derived corpus statistics under CC BY 4.0, with all code under MIT. At the time of writing the benchmark repository ships an MIT \texttt{LICENSE} covering its code only. That data license is applied as part of this release. A reader who checks the repository before the license is added will not find it. We wrote the prompts for this benchmark. Third-party models generated the outputs under each provider's terms, which we read as giving the customer rights in the output. We are not aware of a provider involved claiming copyright in the output. We release the generations under CC BY 4.0 as well. Several countries have not settled who owns machine-generated text. This release is not a warranty on that question. Anyone redistributing the outputs should preserve the model attribution in each record.

We do not redistribute the four human reference corpora, whose terms differ: EnronSent is public domain, the Blog Authorship Corpus is research use only, and Sentiment140 and DISCO carry their own terms. We publish only aggregate statistics derived from them, together with the scripts that rebuild those statistics from local copies obtained under each source's own terms.

\section*{Artifact Availability}

Our benchmark repository \repo{} contains the 112 prompts, the 19,928 raw model outputs, the derived human statistics, the scorer, the site code, and the result logs, all at commit \texttt{f3dc502}. Reproducing the scorecard in this paper needs no code beyond the released repository: \texttt{python3 harness/score.py -{-}run-id full-merged} at that commit reproduces Table~\ref{tab:full}.

Five short scripts ship with the paper for results the scorer does not produce. \texttt{null\_elo.py} reproduces the arena noise floor in Section~\ref{sec:human}, \texttt{mtld\_check.py} the diversity result in Section~\ref{sec:diversity}, \texttt{robustness.py} the opener checks in Section~\ref{sec:opener-robust}, \texttt{constants.py} the eleven constant variants in Section~\ref{sec:ranking}, and \texttt{completecase.py} the missing-data check in Section~\ref{sec:repro}. Also included are the paper source, vector figures, and supporting result CSV files.

\begingroup
\footnotesize
\setlength{\itemsep}{0pt}
\setlength{\parskip}{0pt}
\bibliographystyle{unsrt}
\bibliography{references}
\endgroup

\clearpage
\onecolumn
\raggedbottom
\appendix
\section{Full Mechanical Scorecard}

\begin{table}[H]
\centering
\caption{Full scorecard, sloppiest first. Lower scores mean less measured slop. The interval spans the 2.5th to 97.5th percentiles of each model's rank under a scenario-level cluster bootstrap ($B=500$). We treat models with overlapping intervals as tied. Only Mistral Large is resolved to a point. Rhythm averages the email score with a structural zero from workplace chat, so it is half the email figure (Section~\ref{sec:rhythm}).}
\label{tab:full}
\scriptsize
\setlength{\tabcolsep}{3.6pt}
\begin{tabular}{lrcrrrrrrrrr}
\toprule
& & 95\% rank & \multicolumn{5}{c}{Slop Index by domain} & \multicolumn{4}{c}{Axis mean} \\
\cmidrule(lr){4-8}\cmidrule(lr){9-12}
Model & Rank & interval & Overall & Email & Essay & Chat & Social & Length & Opener & Rhythm & Tells \\
\midrule
Mistral Large & 1 & 1 & 40.6 & 39.9 & 42.6 & 32.1 & 47.8 & 4.7 & 91.2 & 6.1 & 43.2 \\
Claude Fable 5 & 2 & 2--3 & 35.6 & 43.4 & 32.1 & 27.0 & 39.8 & 17.8 & 65.8 & 6.9 & 35.5 \\
GPT-5.6 Terra & 3 & 3--6 & 32.8 & 35.1 & 35.1 & 25.1 & 36.0 & 6.3 & 68.6 & 15.0 & 33.1 \\
Claude Opus 4.8 & 4 & 3--8 & 31.8 & 38.4 & 31.7 & 24.7 & 32.4 & 8.9 & 63.5 & 10.9 & 35.7 \\
GPT-5.6 Sol & 5 & 3--8 & 31.7 & 35.5 & 30.7 & 29.1 & 31.6 & 2.7 & 72.3 & 13.7 & 30.8 \\
GPT-5.6 Luna & 6 & 3--8 & 31.7 & 33.6 & 31.0 & 28.4 & 33.6 & 4.5 & 68.4 & 12.5 & 34.0 \\
GLM 5.2 & 7 & 4--9 & 31.2 & 39.3 & 30.9 & 21.3 & 33.4 & 16.8 & 50.0 & 5.5 & 43.8 \\
Claude Sonnet 5 & 8 & 4--9 & 30.5 & 34.7 & 28.9 & 22.8 & 35.6 & 12.9 & 57.5 & 6.0 & 32.9 \\
Qwen 3.7 Max & 9 & 8--11 & 28.7 & 32.4 & 22.6 & 22.8 & 37.0 & 11.4 & 47.3 & 9.5 & 44.1 \\
Gemini 3.5 Flash & 10 & 9--12 & 27.8 & 39.3 & 18.9 & 21.8 & 30.9 & 16.3 & 41.8 & 10.0 & 40.1 \\
GPT-5.4 Mini & 11 & 9--12 & 26.8 & 30.9 & 28.8 & 20.8 & 26.8 & 3.6 & 56.3 & 18.3 & 27.1 \\
Claude Haiku 4.5 & 12 & 12--16 & 24.9 & 33.6 & 27.4 & 18.1 & 20.6 & 5.7 & 45.9 & 11.4 & 37.4 \\
Gemini 3.1 Pro & 13 & 12--17 & 24.3 & 30.8 & 20.0 & 20.7 & 25.6 & 11.9 & 38.0 & 10.0 & 36.4 \\
Grok 4.5 & 14 & 13--17 & 24.1 & 27.3 & 20.4 & 19.0 & 29.6 & 3.2 & 44.8 & 9.7 & 40.4 \\
DeepSeek V4 Pro & 15 & 8--18 & 23.8 & 20.3 & 20.9 & 13.5 & 40.4 & 19.2 & 24.8 & 1.2 & 43.0 \\
MiniMax M3 & 16 & 13--18 & 23.1 & 30.7 & 20.2 & 15.4 & 26.0 & 15.7 & 28.1 & 6.5 & 41.5 \\
Muse Spark 1.1 & 17 & 12--18 & 23.1 & 35.7 & 15.2 & 18.4 & 23.0 & 9.4 & 39.9 & 18.0 & 27.5 \\
Kimi K2.6 & 18 & 16--18 & 21.1 & 26.9 & 16.7 & 16.1 & 24.6 & 8.0 & 32.2 & 12.6 & 35.5 \\
\bottomrule
\end{tabular}
\end{table}

\section{Scenario and Output Schema}
\label{app:schema}

Each scenario record contains an identifier, domain, category, persona, recipient, context thread, goal, prompt, target length interval, an author note about the intended stress case, a \texttt{split} label, and a flag marking the 20 scenarios used in the pilot. Each output record contains the scenario identifier, domain, category, model alias, sample index, system message, user prompt, raw text, length target, and usage metadata. Each usage block records the route identifier, token counts, reasoning tokens, latency, and estimated cost.

Those 112 scenarios divide into 38 categories, three per category except where noted. Target lengths span 56 distinct bands from 10 to 700 words.

\begin{itemize}[leftmargin=1.1em]
\item \textbf{Email} (30): bad news, cold outreach, decline or pushback, follow-up, internal update, negotiation, scheduling, support reply, thank-you, warm introduction reply.
\item \textbf{Essay} (27): argumentative essay, blog post, commentary, cover letter, how-to blog, newsletter introduction, opinion piece, personal essay, product review.
\item \textbf{Workplace chat} (27): channel announcement, delay to a manager, favor ask, quick question, review request, standup update, urgent reply; plus declining a meeting, group scheduling, and apologizing for a missed message at two each.
\item \textbf{Social} (28): community announcement, Instagram caption, LinkedIn announcement, LinkedIn hiring post, LinkedIn lesson, X post, X reply, X thread; plus launch post at four.
\end{itemize}

\section{Tell Inventory}
\label{app:tells}

The released counter searches for seventeen classes: ritual well-wishing, ``reach out,'' ``circle back,'' not-X-but-Y forms, ``delve,'' ``tapestry'' and related elevated nouns, launch excitement phrases, ``elevate'' and related product language, ``do not hesitate,'' engagement prompts, formal sign-offs, greetings, em-dash glyphs, broad dash use, rule-of-three forms, ``as an AI,'' and apology phrases. Dropping the broad dash counter so that dashes are not weighted twice leaves sixteen scored classes.

The presence of one item does not make a text an example of slop. The axis compares model and corpus rates across many outputs. Table~\ref{tab:tells} reports, for each class and domain, the median score across models. Six classes, including three of the best-known markers, sit at zero in every domain where they are scored at all. Only one class-domain pair falls strictly between the endpoints, so the axis mostly counts saturated markers and its graded range goes almost unused. Items that never separate models from the corpus should be removed. The em-dash class should be reported separately, since it carries the largest gap and the clearest corpus-age confound.

\begin{table}[H]
\centering
\caption{Median score across models for each tell class. $0$ = at or below the human rate; $100$ = at least three times the human rate; $50$ = human rate is zero and the model uses it; $\ast$ = strictly between; ``--'' = neither corpus nor models use it, so the class is dropped from that domain's average.}
\label{tab:tells}
\small
\begin{tabular}{lcccc}
\toprule
Class & Email & Essay & Chat & Social \\
\midrule
Apology & $100$ & $0$ & $100$ & $0$ \\
``As an AI'' & -- & -- & $0$ & -- \\
``Circle back'' & $100$ & $0$ & $0$ & -- \\
``Delve'' & $0$ & $0$ & $0$ & -- \\
``Do not hesitate'' & $0$ & $0$ & $0$ & $0$ \\
Product language & $0$ & $100$ & $0$ & $0$ \\
Em-dash glyph & $50$ & $100$ & $100$ & $50$ \\
Ritual well-wishing & $0$ & $0$ & -- & -- \\
Greeting & $100$ & $100$ & $100$ & $0$ \\
Engagement prompt & -- & $0$ & -- & -- \\
Not-X-but-Y & $0$ & $\ast$ & $0$ & $0$ \\
``Reach out'' & $100$ & $100$ & $0$ & $100$ \\
Rule of three & $0$ & $0$ & -- & $100$ \\
Formal sign-off & $0$ & $100$ & $0$ & $0$ \\
Elevated nouns & $0$ & $0$ & $0$ & $0$ \\
Launch excitement & $0$ & $100$ & -- & $50$ \\
\bottomrule
\end{tabular}
\end{table}

\section{Reproducibility Notes}

We compute every reported mechanical score from the fixed released JSONL file. The repository includes the scenarios, scorer, reference statistics, rebuild scripts, and committed score log. It does not include licensed reference texts, the raw arena vote log, or a complete generation manifest, and Section~\ref{sec:limits} states the resulting limits on reproduction and interpretation.

\end{document}